\pdfoutput=1
\documentclass[11pt,a4paper]{article}
\usepackage[hyperref]{acl2021}
\usepackage{times}
\usepackage{latexsym}

\usepackage{microtype}
\usepackage{multirow}
\usepackage[normalem]{ulem}
\useunder{\uline}{\ul}{}
\usepackage[]{graphicx}
\usepackage{caption}
\usepackage{subcaption}
\usepackage{xcolor}

\aclfinalcopy 

\title{Did the Cat Drink the Coffee?\\Challenging Transformers with Generalized Event Knowledge}

\author{Paolo Pedinotti \\
  University of Pisa \\
  \texttt{paolo.pedinotti@phd.unipi.it} \\\And
  Giulia Rambelli \\
  University of Pisa - Aix-Marseille University \\
  \texttt{\hfill giulia.rambelli@phd.unipi.it} \\\AND
  Emmanuele Chersoni \\
    The Hong Kong Polytechnic University \\
  \texttt{emmanuelechersoni@gmail.com } \\\And
  Enrico Santus \\
  Bayer Pharmaceuticals \\
  \texttt{esantus@gmail.com} \\\AND
  Alessandro Lenci \\
    University of Pisa\\
  \texttt{alessandro.lenci@unipi.it}\\\And
  Philippe Blache\\
  Aix-Marseille University \\
  \texttt{blache@lpl-aix.fr}}

\date{}

\begin{document}
\maketitle
\begin{abstract}
Prior research has explored the ability of computational models to predict a word semantic fit with a given predicate. 
 While much work has been devoted to modeling the typicality relation between verbs and arguments in isolation, in this paper we take a broader perspective by assessing whether and to what extent computational approaches have access to the information about the typicality of entire events and situations described in language (\textit{Generalized Event Knowledge}).

Given the recent success of Transformers Language Models (TLMs), we decided to test them on a benchmark for the \textit{dynamic estimation of thematic fit}. The evaluation of these models was performed in comparison with SDM, a framework specifically designed to integrate events in sentence meaning representations, and we conducted a detailed error analysis to investigate which factors affect their behavior. 
Our results show that TLMs can reach performances that are comparable to those achieved by SDM. However, additional analysis consistently suggests that TLMs do not capture important aspects of event knowledge, and their predictions often depend on surface linguistic features, such as frequent words, collocations and syntactic patterns, thereby showing sub-optimal generalization abilities.

\end{abstract}

\section{Introduction}

People can discriminate between typical (e.g., \textit{A cop arrested a thief}) and atypical events (e.g., \textit{A thief arrested a cop}) and exploit this ability in online sentence processing to anticipate the upcoming linguistic input. Brains have been claimed to be ``prediction machines'' \cite{clark2013whatever} and psycholinguistic research has shown that a crucial ingredient of such predictive ability is the knowledge about events and their typical participants stored in human semantic memory, also referred to as \textbf{Generalized Event Knowledge} (\textsc{GEK}) by \citet{mcrae2009people}. To make an example, if we were asked to think about things that are played with a guitar, we would quickly and more or less unanimously think of words such as \textit{song}, \textit{piece} or \textit{riff}. 

Computational models of predicate-argument typicality, generally referred to as \textit{thematic fit} in the psycholinguistic literature \citep{mcrae1998modeling}, extract typical arguments from parsed corpora. However, \textsc{GEK} is not just storing relations between words: The fact that this knowledge is generalized – that is, it is based on an abstract representation of what is typical – allows us to easily classify new argument combinations as typical or atypical. 
Furthermore, psycholinguistic studies \citep{bicknell2010effects,matsuki2011event} have shown that humans are able to combine and dynamically update their expectations during sentence processing: for example, their expectations given the sequence \textit{The barber cut the \underline{\hspace{0.3cm}}} differ from the ones given \textit{The lumberjack cut the \underline{\hspace{0.3cm}}}, since the integration of knowledge ``cued'' by the agent argument with the verb will lead to the activation of different event scenarios. In Distributional Semantics, sophisticated models of the \textsc{GEK} have been proposed to make predictions on upcoming arguments by integrating the cues coming from the verb and the previously-realized arguments in the sentence \citep{lenci2011composing, chersoni2019structured}. 
Since such knowledge is acquired from both first-hand and linguistic experience \citep{mcrae2009people}, an important assumption of this literature is that, at least for its "linguistic subset", the  \textsc{GEK} can be modeled with distributional information extracted from corpora \citep{chersoni2017logical,chersoni2021not}.

Language Models are trained to make predictions given a context, and thus, they can also be viewed as models of \textsc{GEK}. This approach is promising if one considers the success of recent Transformer-based Language Models (henceforth \textsc{TLMs}), which are trained on huge corpora and contain a massive number of parameters. 
Even if these models receive extensive training and have been shown to capture linguistic properties \citep{jawahar2019does,goldberg2019assessing}, it is not obvious whether they acquire the aspects of \textsc{GEK} that have been modeled explicitly in previous approaches. To the best of our knowledge, Transformers have never been tested on dynamic thematic fit modeling, nor their performance has been compared with traditional distributional models. Our current work is addressing this issue.
\subsection*{Contributions:}
\begin{enumerate}
\item we propose a methodology to adapt \textsc{TLMs} to the dynamic estimation of thematic fit, using a dataset that contains several types of argument combinations differing for their typicality; \item we present a comprehensive evaluation of various \textsc{TLMs} on this task, performed by comparing them to a strong distributional baseline;
\item we conduct further analysis aimed at identifying the potential limitations of \textsc{TLMs} as models of \textsc{GEK}.  
\end{enumerate}

\noindent{}Our results are relevant for researchers interested in assessing the linguistic abilities of \textsc{TLMs}, as well as those working on applications involving \textsc{TLMs}, such as text generation.

\section{Related Work}
In its classical form, the thematic fit estimation task consists in comparing a candidate argument or \textit{filler} (e.g., \textit{wine}) with the typical fillers of a given verb role (e.g., agent, patient, etc.), either in the form of exemplars previously attested in a corpus \citep{erk2007simple,vandekerckhove2009robust,erk2010flexible} or in the form of a vector-based prototype \citep{baroni2010distributional,sayeed2014combining,sayeed2015exploration,greenberg2015verb,greenberg2015improving,sayeed2016thematic,santus2017measuring,chersoni2020word}. Additionally, recent studies explored the use of masked language modeling with BERT for scoring the candidate arguments \citep{metheniti2020relevant}.
Performance in the thematic fit task is typically measured with the correlation between the output scores of the model and human-elicited typicality judgments for verb-argument pairs \citep{mcrae1998modeling,ferretti2001integrating,pado2007integration,zhang2019sp,marton2021thematic}. 

In the simplest and most common version of this task, the typicality of verb argument-pairs is evaluated in isolation. Thematic fit is instead a \emph{dynamic concept}:
The expectations for an argument in a given verb role do not depend just on the verb, but also on the compositional combination with the other arguments in the sentence \citep{bicknell2010effects}.
To check the ability of computational models to account for the compositional update of argument expectations, \citet{lenci2011composing} framed the problem as a binary classification task: A system is presented a sentence pair, with one sentence expressing a typical real-world situation (\textit{The journalist is checking the report}) and the other sentence expressing a plausible but less typical one (\textit{The mechanic is checking the report}), and the task is to assign a higher thematic fit/typicality score to the former. Notice that the two sentences differ only for one argument, and that the ``atypical" one might, however, be a common filler with respect to the verb target role (e.g., \textit{report} is a typical patient for \textit{check}, it is just less plausible in combination with \textit{mechanic} as an agent).

Several models have tried to tackle the ``dynamic" version of the thematic fit task, either based on classical distributional spaces \citep{chersoni2016towards,chersoni2019structured} or on more sophisticated neural network architectures \citep{tilk2016event,hong2018learning}. 
On the evaluation side, those works made use of the experimental materials of the study by \citet{lenci2011composing}, which are, however, limited to agent-verb-patient triples. The recently-introduced DTFit dataset \citep{Vassallo:2018} is, in comparison, larger in size and provides more variety of fillers and roles (including instruments, locations and time).
Other studies introduced larger datasets, but focused on more specific notions of event plausibility (e.g. the plausibility depending on the physical properties of the participants) \citep{wang2018modeling,porada2019can,ko2019linguistically}.

\section{Experimental Settings}
\subsection{Dataset}

The \textbf{DTFit} \citep{Vassallo:2018} dataset
has been specifically designed for the evaluation of dynamic thematic fit. \footnote{All the datasets used for the experiments described in this paper can be found at the link: \url{https://github.com/giuliarambelli/transformers_thematic_fit}.} 
The dataset contains pairs of tuples that differ only for one element, which can be either a typical or atypical filler of a given role in the event described by the tuple (cf. Table \ref{tab:exdtfit}). 
The dataset includes tuples of different lengths, and the typicality of a given argument depends on its interaction with all the other elements. For each tuple, the authors collected typicality judgments by asking English native speakers how common was the event described. Scores range from $1$ (very atypical) to $7$ (very typical). The dataset mainly targets knowledge about professions, but also other typical everyday situations (e.g., what a dog typically eats, what a grandmother typically does). 

The authors created several datasets, which differ with respect to the semantic role of the candidate filler. For our experiments, we selected the datasets created by the authors for the following relations: \{\textbf{Instrument}, \textbf{Time}, \textbf{Location}\}\textsubscript{DTFit}. Additionally, from the original dataset containing agent-verb-patient triples, we derived two datasets, that we named \textbf{Agent}\textsubscript{DTFit} and \textbf{Patient}\textsubscript{DTFit}. In \textbf{Agent}\textsubscript{DTFit}, the tuples forming a pair differ with respect to the typicality of the agent. In \textbf{Patient}\textsubscript{DTFit}, they differ for the typicality of the patient. We thus get a total of five datasets, each of which covers a different semantic relation. The latter two datasets have the same properties of the others, but they put stronger emphasis on
the dynamic nature of thematic fit, as the atypical filler is still a typical complement of the verb alone. Conversely, the atypical candidate fillers in the other datasets are appropriate fillers of the role, but, in most cases, they do not relate to the other elements of the tuple. Therefore, \textbf{Agent}\textsubscript{DTFit} and \textbf{Patient}\textsubscript{DTFit} are more challenging for computational models, as the typicality of a filler can only be determined through the composition of the verb with another argument. Accordingly, models have to update their predictions by accurately taking into account the whole context. 

For each tuple in DTFit, the task for our models is to predict the upcoming argument on the basis of the previous ones. Models were evaluated in terms of Spearman correlation between the human ratings and the models' scores. Moreover, we performed a second evaluation for \textbf{Agent}\textsubscript{DTFit} and \textbf{Patient}\textsubscript{DTFit}, consisting of measuring the accuracy of each system in assigning a higher thematic fit score to typical tuples. 
To the best of our knowledge, the only attempts to test computational models on this dataset have been done by the authors of the original paper and by \newcite{chersoni2019structured}. In both works, distributional prototype models of thematic fit have been used.

\begin{table}
\setlength\tabcolsep{0.5pt}
\centering
\begin{tabular}{llll}
\hline
\textbf{Role} &\textbf{Tuple}      & \textbf{Typical} & \textbf{Atypical}  \\ 
\hline
\hline
\textbf{Agent} & \underline{\hspace{0.3cm}} mix  paint      & painter &  cook  \\

\textbf{Patient}&tailor  sew \underline{\hspace{0.3cm}}     & dress &  wound  \\ 

\textbf{Instrument}&cook	clean	fish  \underline{\hspace{0.3cm}}    & knife &  sponge  \\

\textbf{Time}&cat chase   bird \underline{\hspace{0.3cm}}      & hunting &  marriage  \\ 

\textbf{Location}&sailor	mop	deck \underline{\hspace{0.3cm}}      & boat &  theatre  \\ 
\hline
\end{tabular}
\caption{Examples of tuples from DTFit.}
\label{tab:exdtfit}
\end{table}

\subsection{Models}
In our experiments, we compared the performance of \textsc{TLMs} with the Structured Distributional Model (\textsc{SDM}), which has been recently shown to be an efficient model for the dynamic estimation of thematic fit \cite{chersoni2019structured}.

\subsubsection{Structured Distributional Model}
The \textbf{Structured Distributional Model} (\textsc{SDM}) proposed by \citet{chersoni2019structured} combines word embeddings and formal semantics to specifically represent \textsc{GEK} and the dynamic construction of sentence meaning. Like traditional distributional models of thematic fit, it builds a prototype representation for a given role (e.g., the typical patient of \textit{sing}) from its typical fillers, but its novelty is that the fillers are retrieved from an external resource called \textit{Distributional Event Graph} (henceforth, $DEG$). $DEG$ represents \textsc{GEK} as a graph automatically built from parsed corpora, where the nodes are words associated to a numeric vector, and the edges are labeled with syntactic relations and weighted using statistic association measures. Thus, given a lexical cue $w$, it is possible to identify the events in which $w$ takes part and to retrieve words related to $w$ on both the paradigmatic and the syntagmatic axis.

The formal structure at the basis of \textsc{SDM} consists of two semantic structures: the \textit{linguistic condition} ($LC$), a context-independent tier of meaning that represents the lexical items in a sentence, and the \textit{active context} ($AC$), which accumulates contextual information activated by lexical items. 
The crucial aspect of \textsc{SDM} is that it associates a vector representation to these formal structures: $\vec{LC}$ is the sum of the embeddings of the lexical items of a sentence; $\vec{AC}$, for each syntactic slot, is represented as the centroid vector built out of the role vectors $\vec{r_1},...,\vec{r_n}$ available in $AC$, corresponding to the syntactic associates of the lexical items that have been already processed.

In our implementation of \textsc{SDM}, the $DEG$ was constructed by extracting syntactic relations from a concatenation of the ukWaC corpus \citep{baroni2009wacky}, a dump of Wikipedia 2018 and the British National Corpus \citep{leech1992100}. The final graph contains words with a minimum frequency of 300 and events with a minimum frequency of 30. We used as lexical embeddings the publicly-available FastText vectors extracted from Wikipedia.\footnote{\url{https://fasttext.cc/docs/en/english-vectors.html}}
For our experiments, we built a semantic representation for each tuple in the dataset, like in \newcite{chersoni2019structured}. We used the information in $LC$ and $AC$ to assign a typicality score to each candidate filler of a role in the dataset. The scoring function for a given role filler is the following:
\begin{equation}
 \frac{cos(\vec{f}, \vec{LC}(sent)) + cos(\vec{f}, \vec{AC}(sent))}{2}
\end{equation}

\noindent{}where $\vec{f}$ is the embedding of the candidate filler; $\vec{LC}(sent)$ is a vector obtained from the sum of the embeddings of the verb and of the argument other than $f$; $\vec{AC}$ stands for the updated expectation prototype for the role filled by $f$. In other words, we quantify the typicality of an argument given a tuple as the average of i.) the cosine similarity between the argument embedding and the additive combination of the other argument vectors ($\vec{LC}$), and ii.) the cosine similarity between the argument embedding and the prototype vector representing the active context ($\vec{AC}$).
In the cases where $\vec{AC}$ cannot be derived (because $DEG$ does not store syntactic relations involving the context words), we take only the cosine between $\vec{f}$ and $\vec{LC}(sent)$ as the final score.

\subsubsection{Transformer-based Language Models}
We experimented with four \textsc{TLMs} to test how different architectures, training objectives, and sizes of the training corpus affect performance.\footnote{For all experiments involving \textsc{TLMs}, we use pre-trained models available in the HuggingFace's Python library Transformers \citep{wolf2019huggingface}.}

\textbf{BERT} \cite{Devlin:etal:2019} consists of a series of stacked Transformer encoders. It was trained using both a masked language modeling objective (i.e., predicting a masked word from its left- and right-context), and a next sentence prediction objective (i.e., whether a sentence follows another sentence or not), on a combination of the BooksCorpus and English Wikipedia (13GB in total). The model uses WordPiece vocabulary. To test if the model size can affect BERT performance, we used both the \texttt{base} (Number of layers=12, Hidden size=768) and the \texttt{large} (L=24, H=1024) versions. 

\textbf{RoBERTa} \citep{liu2019roberta}, which we used in the \texttt{large} version, is based on the same architecture as BERT, but it was trained on a much larger corpus (160GB) and without the next sentence prediction objective. In our experiments, we used the large version (L=24, H=1024). 

In contrast with the bidirectional nature of BERT and RoBERTa, \textbf{GPT2} \cite{radford2019gpt2} is a uni-directional \textsc{LM}, which means that the training objective is to predict the next word, given all of the previous ones. It was trained on WebText, for a total of 8 million documents of data (40 GB). We employed the \texttt{medium} version of GPT2 (L=24, H=1024). We chose GPT2-medium since its dimensions are comparable to those of BERT and RoBERTa large. Moreover, both RoBERTa and GPT2 make use of a Byte-Pair Encoding tokenizer.  

For our investigation, we designed the experiment as follows. First, we derived simple sentences from the tuples by adding definite articles to the words, [CLS] at the beginning of the input and a period to signal the end of the sentence (e.g., \texttt{[CLS] The tailor sewed the dress.}). Then, we masked the candidate filler (\textit{dress} in the example) and we computed the probability distribution of the entire model's vocabulary for that position.
The model typicality score is the probability assigned to the candidate filler, when the candidate filler is included in the model’s vocabulary. In case a word to be scored is not included in the vocabularies of all the models that we used, we decided to disregard its tuple and the respective typical/atypical counterpart. For this reason, the final results only take in consideration a subset of the original datasets, which varies from model to model.
Additionally, we computed a baseline for each Transformer model, where the model is prevented from attending to the other tokens in the sequence when making predictions. 

\section{Results and Analysis}

\begin{table*}[h!tb]
\setlength\tabcolsep{3.7pt}
\centering
\small
\begin{tabular}{l@{}rccccc}
\hline
     & Coverage       & SDM           & BERT-base(line) & BERT-large    & ROBERTA-large & GPT-2 medium  \\
\hline \hline
\textbf{Agent}\textsubscript{DTFit}  & 105/134   & 0.58          & 0.46  (0.1)    & 0.53          & \textbf{0.64} & -             \\
\textbf{Patient}\textsubscript{DTFit} &   323/402  & 0.62          & 0.59  (0.06)    & \textbf{0.64} & \textbf{0.64} & 0.63          \\
\textbf{Instrument}\textsubscript{DTFit} & 31/100 & \textbf{0.58} & 0.52  (0.08)    & 0.53          & 0.5           & 0.5           \\
\textbf{Time}\textsubscript{DTFit} &    89/100    & 0.58          & 0.63  (0.06)    & 0.64          & \textbf{0.66} & \textbf{0.66} \\
\textbf{Location}\textsubscript{DTFit}&  115/150  & 0.65          & 0.72  (0.06)    & 0.71          & 0.73          & \textbf{0.74}\\
\hline
\end{tabular}
\caption{Spearman Correlation for the DTFit datasets.}
\label{tab:results_DTFit}
\end{table*}

\begin{figure*}[t!hb]
\centering
\begin{subfigure}{.4\textwidth}
  \centering
  \includegraphics[width=.8\linewidth]{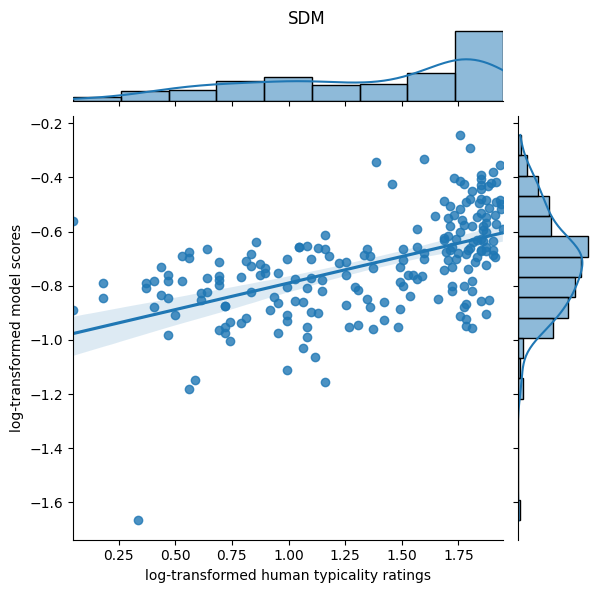}
  \caption{}
  \label{fig:sdm-agent}
\end{subfigure}%
\begin{subfigure}{.4\textwidth}
  \centering
  \includegraphics[width=.8\linewidth]{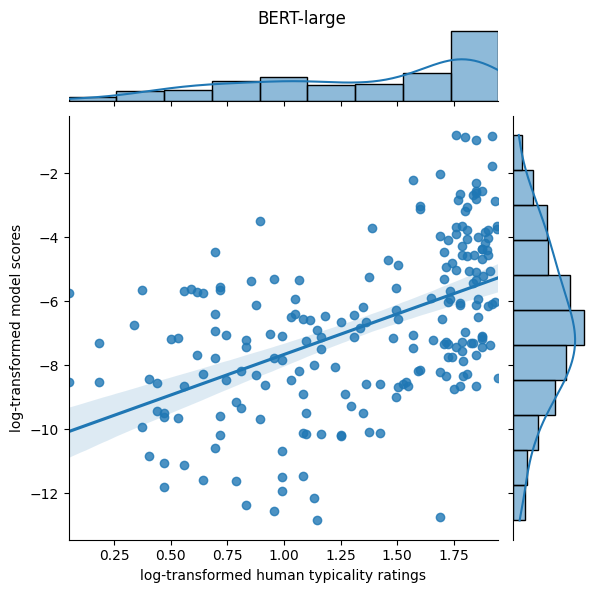}
  \caption{}
  \label{fig:bert-agent}

\end{subfigure}
 
\begin{subfigure}{.4\textwidth}
  \centering
  \includegraphics[width=.8\linewidth]{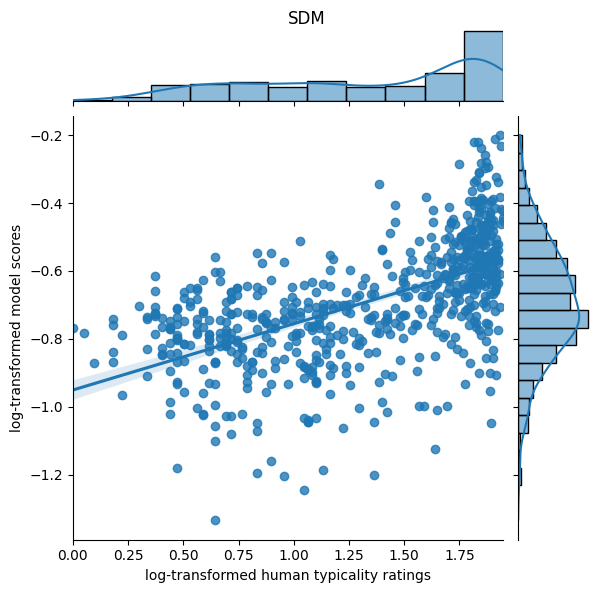}
  \caption{}
  \label{fig:sdm-patient}
\end{subfigure}%
\begin{subfigure}{.4\textwidth}
  \centering
  \includegraphics[width=.8\linewidth]{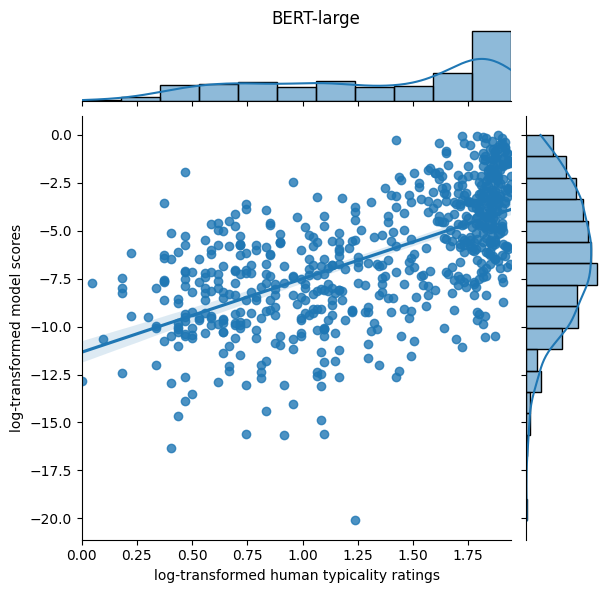}
  \caption{}
  \label{fig:bert-patient}
\end{subfigure}%
\caption{Correlation of elicited judgments and model-derived scores for \textbf{Agent}\textsubscript{DTFit} (a-b) and \textbf{Patient}\textsubscript{DTFit} (c-d) datasets.}
\label{fig:correls}
\end{figure*}

In this section, we provide the results of the experiments on the DTFit datasets.
Since the models cover different portions of the original tuples, we performed the evaluation over the common pairs.

Table \ref{tab:results_DTFit} reports the correlation scores for all the five datasets.\footnote{We do not computed GPT-2 scores for \textbf{Agent}\textsubscript{DTFit}, as the model cannot make predictions based on context because the candidate filler occurs at the beginning of the sentence.} Values in brackets refer to the Spearman correlation obtained by the baseline. As the baseline scores are very similar across models, we reported the results only for BERT-base.

At a glance, we observe that both \textsc{SDM} and \textsc{TLMs} obtain quite strong correlations, going from $0.46$ to a maximum of $0.74$ across datasets and models. 
Specifically, we notice that \textsc{TLMs} tend to reach higher performances compared to the distributional approach. However, a marginally significant improvement of the correlations over \textsc{SDM} is obtained only for \textbf{Location}\textsubscript{DTFit} ($p<0.05$ for Locations, $p<0.1$ for the other roles).\footnote{The $p$-value was computed with Fisher’s r-to-z transformation, one-tailed test.}
This result is interesting, considering that \textsc{SDM} is trained on a really small corpus compared to \textsc{TLMs} (for instance, RoBERTa is trained on 160 GB of text).
Another remark is that even if \textsc{TLMs} differ for architecture, training objective and data, BERT-large, RoBERTa and GPT-2 tend to achieve very similar performances, while correlation scores are lower for BERT-base. 

As there is no significant difference between \textsc{SDM} and \textsc{TLMs} results, we plotted an example of the relationship between the human ratings and the model-derived scores to provide a better picture of the models' predictions. For visualization purposes, we applied a logarithmic transformation to the scores. For \textbf{Agent}\textsubscript{DTFit}, we observe that \textsc{SDM} and BERT-large have a different trend. In the former (see Figure \ref{fig:sdm-agent}), the majority of the points follow a roughly linear relationship, and there is a small variation around the regression line (with few outliers). On the contrary, BERT-large scores show more variance (Figure \ref{fig:bert-agent}).
This trend is confirmed (even if it is less evident) for \textbf{Patient}\textsubscript{DTFit}, where both \textsc{SDM} (Figure \ref{fig:sdm-patient}) and BERT-large (Figure \ref{fig:bert-patient}) have a large amount of variance, and quite a few outliers. 
To verify these observations, we compared the sum of the BERT-large residuals with that of \textsc{SDM} (we first normalized the models' scores with min-max scaling in order to make them comparable). For both subsets, the sum of residuals is higher for BERT-large than \textsc{SDM}, which is especially the case for \textbf{Agent}\textsubscript{DTFit} ($31.43$ versus $17.85$; $67.04$ versus $63.47$ for \textbf{Patient}\textsubscript{DTFit}). 

Finally, we also performed a \textit{binary classification task} for \textbf{Agent}\textsubscript{DTFit} and \textbf{Patient}\textsubscript{DTFit}. In this case, we evaluated models on their ability to assign a higher score to the filler in the typical condition.
As shown in Table \ref{tab:accs} (left columns), the accuracy values are always high and the \textsc{TLMs} scores are comparable with the \textsc{SDM} ones. 

\section{Do Transformers Really Encode \textsc{GEK}?}
\label{sec:other-expe}
The above results \emph{prima facie} suggest that \textsc{TLMs} are able to model the dynamic interaction between the sentence elements to compute the typicality value of a candidate filler. However, analyzing the errors of the \textsc{TLMs} can be revealing of how they make their predictions.

Table \ref{tab:exberterrors} presents some of the \textbf{Patient}\textsubscript{DTFit} pairs where BERT-base prefers the atypical filler. In all these cases, BERT simply seems to rely on frequent verb objects, without composing and integrating the verb expectations with information from other elements of the context (the agent in this case), which is a key aspect of human \textsc{GEK} and is reflected in the typicality judgments.
However, we cannot make any claims about the event knowledge of \textsc{TLMs} from these examples alone, as only in some cases (such as \textit{The cat drank the \textbf{coffee}}) the atypical tuples evoke events unlikely to take place in the real world (i.e., it may happen frequently that \textit{a chemist pours the juice}, even if this is not a typical action for a chemist). To better understand if this can lead \textsc{TLMs} to make really implausible predictions, we carried out an additional experiment where we tested the models on a diagnostic dataset controlled for the frequency of the association between the verb and the filler. In this experiment, we also tried to address the question of whether \textsc{TLMs} rely more heavily on the local context when making predictions.


Furthermore, \textsc{TLMs}' natural preference for what is more frequent could help them in the typicality task, as a typical event is often a frequent one. Their good performance could be due to the fact that they memorize frequent sequences during training. Therefore we tested \textsc{TLMs} on a different dataset, in which atypical but physically plausible events (e.g., \textit{The cloth erased the cream}) are distinguished from atypical and implausible ones (e.g., \textit{The cloth erased the house}). Frequency effects on performance should be alleviated in this setting, as both types of events in the dataset are atypical and, hence, rare. This task requires fine-grained knowledge of the properties of arguments, which is still an important component of \textsc{GEK}.

Additionally, different frequency variations in the training data could influence \textsc{TLMs} performance. Since the models’ knowledge of the world is mediated by language, it is likely that an argument filler may or may not be predicted depending on the frequency of the word chosen to refer to it. We investigated this issue by testing the models on another diagnostic dataset obtained by replacing typical fillers with low-frequency synonyms.

The last question we explored is whether \textsc{TLMs} can be influenced by the way statements of event typicality are syntactically expressed. So, we evaluated \textsc{TLMs} by feeding them with sentences encoding typical events with a transformed and more complex syntactic form than the one used in the DTFit experiments. 

\setlength\tabcolsep{1pt}
\begin{table}
\small
\centering
\begin{tabular}{lcc|cc}
\hline
                 & \multicolumn{2}{c}{\textbf{DTFit}}          & \multicolumn{2}{c}{\textbf{Wang2018}}         \\
                  & Agent        & Patient        & Agent        & Patient           \\ \hline \hline
SDM                  &\textbf{.89}      & \textbf{.91}     & .65    & .66    \\
BERT-base              & .77      & .85     &.76   & .63      \\
BERT-large                   & .83 & .89 & \textbf{.77}  & .65\\
ROBERTA-large                  & \textbf{.89} & \textbf{.91} & .76  & \textbf{.73 }  \\
GPT-2 medium                 & - & .90 & - & .64  \\
\hline     
\end{tabular}
\caption{Accuracy in the binary classification task for DTFit (agent and patient roles) and Wang2018 datasets.}
\label{tab:accs}
\end{table}

\begin{table*}
\centering
\small
\begin{tabular}{lll}
\hline
\textbf{Tuple}      & \textbf{Expected} & \textbf{Preferred}  \\ 
\hline
\hline
mason  mix \underline{\hspace{0.3cm}}     & cement (H=6.65, M=-8.41) &  soup (H=1.95, M=-5.54)  \\

climber  climb \underline{\hspace{0.3cm}}     & rock (H=6.8, M=-5.29) &  staircase (H=5.6, M=-4.05)  \\ 

blacksmith  pour \underline{\hspace{0.3cm}}     & metal (H=6.5, M=-4.03) &  wine (H=1.6, M=-1.6)  \\ 

chemist  pour \underline{\hspace{0.3cm}}     & compound (H=6.25, M=-8.4) &  juice (H=2.75, M=-5.18)  \\ 

cat  drink \underline{\hspace{0.3cm}}     & milk (H=5.6, M=-2.89) &  coffee (H=1.45, M=-3.65) \\

\hline
\end{tabular}
\caption{Examples of errors (BERT-base, Patient\textsubscript{DTFit}). H= Human score, M=Model's log probability.}
\label{tab:exberterrors}
\end{table*}

\paragraph{I. \textsc{TLMs} seem to prefer frequent collocations, but only when they are plausible.} 
Errors reported in Table \ref{tab:exberterrors} suggest the tendency of \textsc{TLMs} to predict frequent complements of the verbs, irrespective of whether they are coherent with the rest of the tuple. We questioned to what extent salient local word co-occurrences make the models ``blind" to the rest of the context and thus compromise the plausibility of their predictions.
To investigate this behavior, we generated a new diagnostic dataset. 
The dataset is a small ($31$ pairs) subset of \textbf{Patient}\textsubscript{DTFit}, where the atypical filler in each pair was replaced with another noun that has a very strong association with the verb in the tuple. We computed the association between the verb and its direct objects using Local Mutual Information (LMI) \cite{evert2008corpora}. Since LMI is computed by multiplying the Pointwise Mutual Information and the frequency of the two words in a grammatical relation, it assigns higher values to combinations that are both common and informative. We chose the new atypical fillers among the words with the highest LMIs. We chose words that give rise to odd events when integrated with the rest of the context.
To approximate the word distributions encountered in the training data, we extracted LMI values from a 2018 dump of English Wikipedia and we evaluated only the BERT model (base and large) on the new dataset, as Wikipedia is a considerable part of the training only for this model. 
Examples of the new test pairs are the following: \textit{The terrorist released the \textbf{hostage}/ \textbf{album}}, \textit{The truck hit the \textbf{car}/ \textbf{ball}}, \textit{The soldier heard the \textbf{command}/ \textbf{case}}. 

To evaluate BERT performance, we computed the accuracy scores on the diagnostics dataset in the same way as in the main experiment (binary classification task). Results show that the models generally assign low probabilities to atypical fillers. They choose the atypical event in some cases ($9$ in BERT-base, $6$ in large), but mainly when the contrast between the atypical event and our expectations is less evident (\textit{The smuggler sold the \textbf{property}} is preferred to \textbf{weapon}, \textit{The soldier throw the \textbf{ball}} is preferred to \textbf{bomb}).

As already observed in the main experiment, BERT seems to be able to look beyond salient local associations and build representations of global events flexibly.
However, this issue should be further explored for the other roles as well. For instance, given the sentence \textit{The engineer completed the project in the \underline{\hspace{0.3cm}}}, the models must consider more contextual elements to make the correct prediction. 

On the other hand, even if \textsc{SDM} design aims at capturing this aspect of \textsc{GEK}, the manipulations we made in this dataset cause a drop in the model performance ($14$ pairs out of $31$ are classified wrongly). This drop is probably due to aspects of the implementation such as data sparsity. Specifically, if there are no events in which the subject occurs with a direct object, the prototype of the patient is built only from the verb's most associated patients, disregarding the fact they are implausible given the whole context.

\paragraph{II.  \textsc{TLMs} know more about what is typical than what is possible.} The use of typicality datasets such as DTFit for the estimation of the models’ \textsc{GEK} has some limitations. \textsc{TLMs}' ability to reproduce combinations encountered frequently during training could be the reason for high performances in the typicality task, since what is most typical often occurs most frequently. However, \textsc{GEK} is not just memory of exemplars, but it requires fine-grained knowledge of the properties of objects and it involves reasoning processes such as abstraction and comparison between objects and prototypical concepts. 

To evaluate \textsc{TLMs} on a setting where frequency variations in the training corpus have a minor impact, we used the dataset realized by \citet{wang2018modeling} (henceforth, \textbf{Wang2018}). This dataset represents a benchmark for the task of semantic physical plausibility \citep{bagherinezhad2016are}, that is, distinguishing an atypical but physically plausible event such as \textit{The student climbed the ship} from an atypical and physically implausible one such as \textit{The student climbed the water}. The dataset contains agent-verb-patient (SVO) triples divided into plausible and implausible.
From the original dataset, which contains $1,540$ plausible and $1,540$ implausible triples, we derived two subsets containing pairs of triples differing either for the agent or for the patient role filler (obtaining $222$ and $394$ pairs respectively).

Table \ref{tab:accs} reports the resulting accuracy values. In general, the models' scores are lower than in the typicality task (min. $0.64$, max. $0.77$), and in some cases they are not much higher than random performance. Moreover, in many cases the models could be facilitated by the existence of an association between the plausible filler and the verb of the event, as in \textit{The ant \textbf{built} the \textbf{wall}} and in \textit{The chair \textbf{absorbed} the \textbf{water}}. 
Nevertheless, the results demonstrate that the notion of plausibility is harder to model compared to typicality, and invite caution when making claims about \textsc{TLMs} world and event knowledge. In fact, the results suggest that even if it were true that \textsc{TLMs} develop some generalization skills from training, they still miss many predictions about possible events, which instead humans easily make on the basis of their commonsense knowledge. 

This dataset is also difficult for \textsc{SDM}, which obtains scores lower than those of the \textsc{TLMs} ($0.65$ for Agent and $0.66$ for Patient). Even if \textsc{SDM} should be better at reproducing generalization through the construction of prototypical fillers, the model's distributional representations seem to fail to capture the specific properties that are relevant for the dataset items, namely physical properties of objects (liquid-solid, large-small, etc.). The lack of such properties constitutes a limitation of distributional models of word meaning based on text data only, which is why, in previous studies, world knowledge was explicitly injected into the models for the physical plausibility task \cite{wang2018modeling}.

\paragraph{III. \textsc{TLMs} do not extend fit judgments to low frequency synonyms.} To test whether \textsc{TLMs} consider an entity more or less likely to take part in an event depending on the word used to refer to that entity, we evaluated them on a new diagnostic dataset of $39$ pairs, generated from a subset of \textbf{Patient}\textsubscript{DTFit}. In this setting, the typical filler in each pair was replaced with a low-frequency word that is semantically related to the original one. To choose an appropriate substitute, we first extracted a set of synonyms according to two lexical resources (WordNet, Lexico.com). Then, we picked a word that 1) is less frequent than the original filler and 2) has a frequency lower than $300,000$.
For the same reasons described in the first additional experiment, we extracted statistics from a 2018 dump of English Wikipedia and evaluated only BERT on the new dataset. Examples of substitutions are the following: \textit{The botanist examined the \textbf{plant}} $\rightarrow$ \textit{\textbf{flora}}, \textit{The waiter cleared the \textbf{restaurant}} $\rightarrow$ \textit{\textbf{tavern}}, 
\textit{The veterinarian examined the \textbf{dog}} $\rightarrow$ \textit{\textbf{puppy}}.
It is interesting to observe that these variations pose serious difficulties to the models, as their accuracy on the diagnostics dataset is close or lower to the random level (BERT-base: $0.37$, BERT-large: $0.53$). For example, BERT considers \textit{The terrorist released the \textbf{captive}} as less probable than \textit{The terrorist released the /\textbf{book}}, and the same occurs for \textit{The mother prepared the \textbf{provisions}/\textbf{gun}}, and \textit{The carver built the \textbf{bust}/\textbf{house}}. 

These results cast doubts that current \textsc{TLMs} can constitute plausible models of event knowledge: they tend to reproduce the patterns that are frequently observed in the data, and their good performance is disrupted once these are replaced with semantically equivalent, but less frequent ones. This means that they lack the abstract semantic knowledge of human subjects, whose predictions are more flexible thanks to inference mechanisms such as generalization to concepts sharing semantic features. 
At least in principle, models aiming at building prototypes of ideal role fillers (such as the distributional models of thematic fit) are more cognitively realistic, since they are less dependent on specific words. However, they may still show sub-optimal performance in this diagnostic dataset as they are based on the quality of the distributional representations, which is lower for words that have low frequency in corpora. This is confirmed by the performance of \textsc{SDM} on the dataset (the accuracy is $0.51$). 

\begin{table}[h!tb]
\centering
\small
\begin{tabular}{lccc}
\hline
                    & transitive & cleft & wh-interrogative\\
                    \hline \hline
\textbf{Agent}\textsubscript{DTFit}      & 0.64       & {\ul -0.13} & 0.62        \\
\textbf{Patient}\textsubscript{DTFit}   & 0.64       & {\ul 0.26 }  & 0.51        \\
\textbf{Instrument}\textsubscript{DTFit} & 0.5        & {\ul 0.10 } & 0.6         \\
\textbf{Time}\textsubscript{DTFit}       & 0.66       & 0.33  & 0.64        \\
\textbf{Location}\textsubscript{DTFit}   & 0.73       & 0.67  & 0.73      \\
\hline
\end{tabular}
\caption{Spearman Correlation for DTFit datasets using RoBERTa-large and input sentences with different word orders.}
\label{tab:word-order}

\end{table}

\paragraph{IV. \textsc{TLMs} can be influenced by the surface structure of sentences}
Finally, we analyzed to what extent \textsc{TLMs}' ability to predict the fit of a word in a given role arises from the observation of recurrent word order patterns during pre-training (e.g., the fact that an actor's award-winning event is canonically expressed with active sentences, in which \textit{award} follows the words \textit{actor} and \textit{won}), rather than being based on a deep understanding of the semantic relations between the words. 

To explore this issue, we modified DTFit tuples to create two different versions of the dataset, each with examples of a syntactic construction different from the English canonical word order. 
Specifically, we experimented with \textbf{cleft} (\textit{It was the {\ul award} that the actor won}, \textit{It was on the {\ul ring} that the boxer delivered the punch}) and \textbf{wh-interrogative} sentences (\textit{Which award did the actor win?}, \textit{On which {\ul ring} did the boxer deliver the punch?}).

We evaluated this new set of sentences using RoBERTa-large (cf. Table \ref{tab:word-order}). We observe that the model is not particularly affected by the interrogative structure. Conversely, the model suffers from the cleft construction for all semantic roles except for Location ($\rho$=$0.67$). If we ask the model to generate the most likely words to appear in that position, we observe that word predictions in the new construction are more general and less dependent on the \textsc{GEK} associated with the other words in the sentence, 
proving that \textsc{TLMs} are affected by the surface syntactic shape of the linguistic input, since the cleft construction is less frequent and presents a less canonical word order. 
For instance, given the sentence \textit{It was with the [MASK] that the guard opened the door}, RoBERTa generates the following possible fillers: \textit{gun} (P=$0.044$), \textit{crowd} (P=$0.020$), \textit{sword} (P=$0.016$), and then \textit{key} (P=$0.016$), while in the active sentence \textit{key} is correctly predicted as the most probable word (P=$0.22$). In this specific case, it seems that the model only looks at the word nearby (\textit{guard}) to make a prediction, disregarding the entire context.
Generally, the agent role shows the worst results, obtaining $-0.13$. Note that \textsc{SDM} is not affected by these variations by design, since its predictions are based on semantic roles derived from the syntactic analysis of the sentence, which is explicitly provided to the model.

\section{Conclusions}


In this paper, we tested Transformer-based Language Models on tasks related to Generalized Event Knowledge.
In the main experiment, we evaluated their ability to model event typicality, that is, discern typical from atypical events, on a dataset designed for this task, DTFit. Results show that \textsc{TLMs} scores positively correlate with human judgments. However, they do not significantly outperform the distributional prototype-based model (\textsc{SDM}) that we selected for comparison. This confirms the ability of \textsc{SDM} to dynamically update the semantic representation of a sentence, which was recently shown for the challenging task of logical metonymy interpretation \citep{rambelli2020comparing}.

However, we decided to go beyond the simple evaluation against human judgments. We carried out several additional small-scale experiments with the specific aim to understand which factors could affect the predictions of \textsc{TLMs}. The results suggest that models are often too dependent on what they observe during training and lack some key aspects of human event knowledge.
 In particular, we observed that, in some cases, they are unable to compose all elements of the input to make predictions, and they tend to rely more on salient local associations between words. However, further analysis is needed. Secondly, their performance drop on the physical plausibility task, which requires the ability to infer physical properties necessary for an object to participate in a given event.
 Lastly, their probabilities are dependent on the specific words that have to be predicted rather than on their meaning, and on the canonical word order in which these words tend to occur. Noticeably, even a distributional model of event knowledge (\textsc{SDM}) showed similar limitations, generally likely to be due to data sparsity and inherent limitations of distributional representations obtained from text data.

To conclude, we believe that the experiments we reported are the first step towards a deep investigation of ``how general'' is the Generalized Event Knowledge in computational models. Future work might include the creation of a larger version of our diagnostic datasets, in order to make available to NLP researchers a more robust benchmark for tasks related to Generalized Event Knowledge.



\section*{Acknowledgements}
This work, carried out within the Institut Convergence ILCB (ANR-16-CONV-0002), has benefited from support from the French government, managed by the French National Agency for Research (ANR) and the Excellence Initiative of Aix-Marseille University (A*MIDEX).
We thank the anonymous reviewers for their insightful feedback.

\bibliographystyle{acl_natbib}
\bibliography{anthology,acl2020,acl2021}


\end{document}